\documentclass[10pt,twocolumn,letterpaper]{article}

\usepackage{iccv}
\usepackage{times}
\usepackage{epsfig}
\usepackage{graphicx}
\usepackage{amsmath}
\usepackage{amssymb}

\graphicspath{{./figures/}}
\usepackage{graphicx}
\usepackage{subfig}
\usepackage{color,soul}
\usepackage{array}
\usepackage{multirow}
\usepackage{color}	
\usepackage{url}
\usepackage[bookmarks=false]{hyperref}	
\usepackage[inline]{enumitem}
\usepackage{xcolor}
\usepackage{svg}

\usepackage{authblk}
\usepackage[font=small]{caption}
\iccvfinalcopy 

\def\iccvPaperID{515} 

\ificcvfinal\fi
\begin{document}

\title{WeText: Scene Text Detection under Weak Supervision}

\author[1]{Shangxuan Tian} 
\author[2]{Shijian Lu}
\author[3]{Chongshou Li}
\affil[1]{Visual Computing Department, Institute for Infocomm Research \tabularnewline
}
\affil[2]{ School of Computer Science and Engineering\\ Nanyang Technological University \tabularnewline
}
\affil[3]{Department of Industrial Systems Engineering and Management \\ National University of Singapore  \tabularnewline
	\url{tianshangxuan@u.nus.edu, Shijian.Lu@ntu.edu.sg, iselc@nus.edu.sg} 
}

\maketitle

\begin{abstract}
   
   The requiring of large amounts of annotated training data has become a common constraint on various deep learning systems. In this paper, we propose a weakly supervised scene text detection method (WeText) that trains robust and accurate scene text detection models by learning from unannotated or weakly annotated data. With a ``light'' supervised model trained on a small fully annotated dataset, we explore semi-supervised and weakly supervised learning on a large unannotated dataset and a large weakly annotated dataset, respectively. For the unsupervised learning, the light supervised model is applied to the unannotated dataset to search for more character training samples, which are further combined with the small annotated dataset to retrain a superior character detection model. For the weakly supervised learning, the character searching is guided by high-level annotations of words/text lines that are widely available and also much easier to prepare. In addition, we design an unified scene character detector by adapting regression based deep networks, which greatly relieves the error accumulation issue that widely exists in most traditional approaches. Extensive experiments across different unannotated and weakly annotated datasets show that the scene text detection performance can be clearly boosted under both scenarios, where the weakly supervised learning can achieve the state-of-the-art performance by using only 229 fully annotated scene text images. 
   
\end{abstract}

\section{Introduction}

Automatic reading texts in scene images has attracted growing interest in recent years due to its great advantages in image content understanding and contextual information inference. It has been widely used in various tasks such as multilingual image translation (Google translate \cite{lancehandson}), vehicle auto-navigation \cite{smith2016end}, object recognition \cite{karaoglu2012object}, and assistive smartphone applications for visually impaired people \cite{looktel}. An indispensable component of an automatic scene text reading system is scene text detection under unconstrained conditions. This is still a very open research challenge due to the tremendous complexity imposed by diverse text fonts and styles, arbitrary text sizes, various geometric distortions, complex image backgrounds, uncontrolled illuminations, \etc.

\begin{figure}[!t]
	\centering	
	\includegraphics[width=0.45\linewidth, height=0.12\textheight]{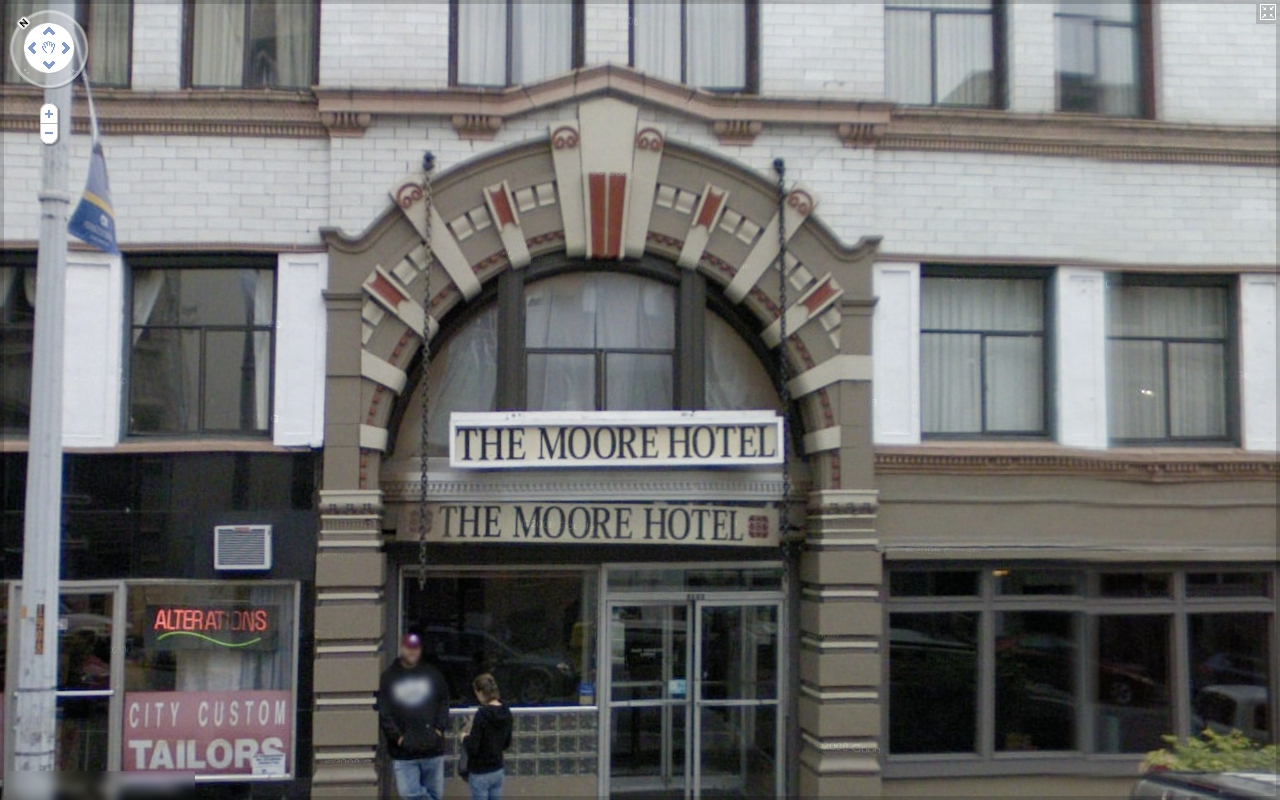}
	\includegraphics[width=0.45\linewidth, height=0.12\textheight]{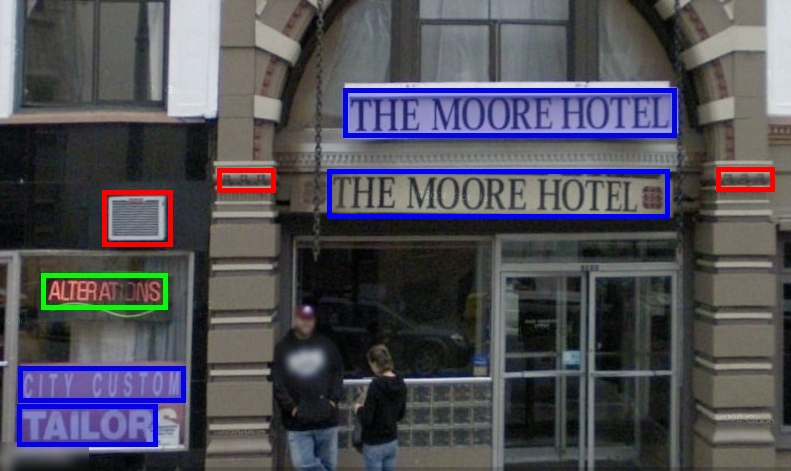} \\ \vspace{1mm}
	\includegraphics[width=0.45\linewidth, height=0.12\textheight]{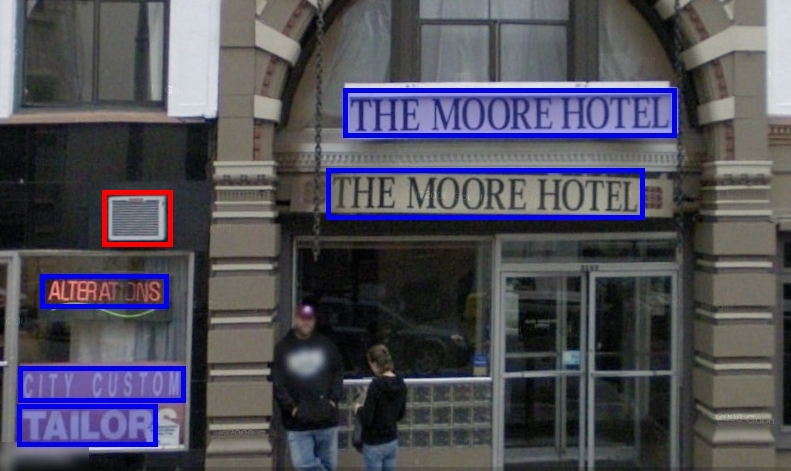}
	\includegraphics[width=0.45\linewidth, height=0.12\textheight]{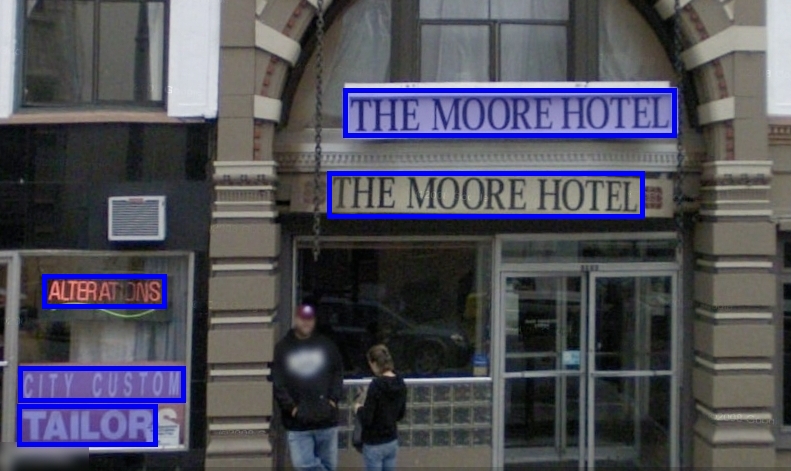}
		\caption{Text detection examples of the proposed WeText system. In the top row from left to right are one sample image and detection outputs using the baseline model. Images in the bottom row from left to right are text detections using the proposed semi-supervised and weakly supervised learning approaches, respectively. Blue boxes indicate correct detections while red boxes and green boxes indicate false positives and false negatives, respectively. Detection results have been zoomed in for better visualization.}
	\label{Fig:show_result_intro}
	\vspace{3mm}
\end{figure}

Two approaches have been explored to address the scene text detection challenge. The first is character based, which first detects character candidates by particular operators such as Stroke Width Transform (SWT) \cite{epshtein2010detecting}, Maximally Stable Extremal Regions (MSER) \cite{neumann2012real, yin2014robust}, sliding windows \cite{tian2015text}, followed by identifying the real characters with a pre-trained text/non-text classifier. Words or text lines are further determined by grouping the detected characters with heuristic rules \cite{huang2014robust, yin2014robust} or more sophisticated graph models \cite{tian2015text, yin2015multi}. The second approach is to detect words directly either by generating word proposals \cite{jaderberg2016reading, zhong2016deeptext} or regressing word bounding boxes from default anchor boxes \cite{liao2016textboxes}. This approach is simpler and more efficient compared with the character-based approach. On the other hand, it does not work well with multi-oriented text as word proposals tend to detect horizontal texts. In addition, many non-Latin languages such as Chinese do not have a clear word boundary which greatly restricts its applicability.

We take a character based approach due to its flexibility in dealing with multilingual and multi-oriented texts in scenes. However, the character based approach has two major constraints. First, a robust and accurate character detector requires a large amount of annotated character images that are time consuming and costly to prepare. 
Second, the current character based approach, which first detects characters candidates and then identifies true characters by a text/non-text classifier, is complicated and also accumulates errors.

In this paper, we propose a weakly supervised scene text detection framework (WeText) that is capable of learning a robust and accurate scene text detector with a small amount of annotated character images. The idea is to first train a ``light" supervised model by using a small amount of fully annotated character images and then apply the model on a large amount of unannotated or weakly annotated images to search for positive training samples. The searched samples are combined with the small amount of annotated images to re-train a more robust and accurate detector. We investigated two learning strategies including semi-supervised learning that requires no annotations and weakly supervised learning where the character searching is guided by high-level annotations of words or text lines. In addition, we adapt regression based deep networks and design a proposal-free character detector that integrates the character candidate detection and text/non-text classification into a single process to reduce the error accumulation. To the best of our knowledge, this is the first work that uses regression based deep networks for scene character detection, and it is also the first work that fully studies the impact of weakly supervised learning for scene text detection. Example results of the proposed WeText framework are given in Figure \ref{Fig:show_result_intro}.

The contributions of our work are twofold. First, we propose a weakly supervised framework that trains a robust and accurate scene text detector by using unannotated or weakly annotated data. The proposed framework aims to address the data annotation constraint faced by many deep learning systems. In particular, it exploits word-level annotations to guide the search for character-level training samples, and the benefits are demonstrated by the great performance gains in scene text detection. Second, we design a proposal-free scene character detector which directly predicts character bounding boxes and text confidence without the complicated candidate detection and classification processes. The integrated detection approach solves the error accumulation issue and greatly improves the accuracy and efficiency for scene text detection. Experiments show that our proposed weakly supervised model can achieve state-of-the-art performance using only 229 images with character annotations. 

\begin{figure*}[!t]
	\centering		
	\includegraphics[width=0.7\linewidth, height=0.24\textheight]{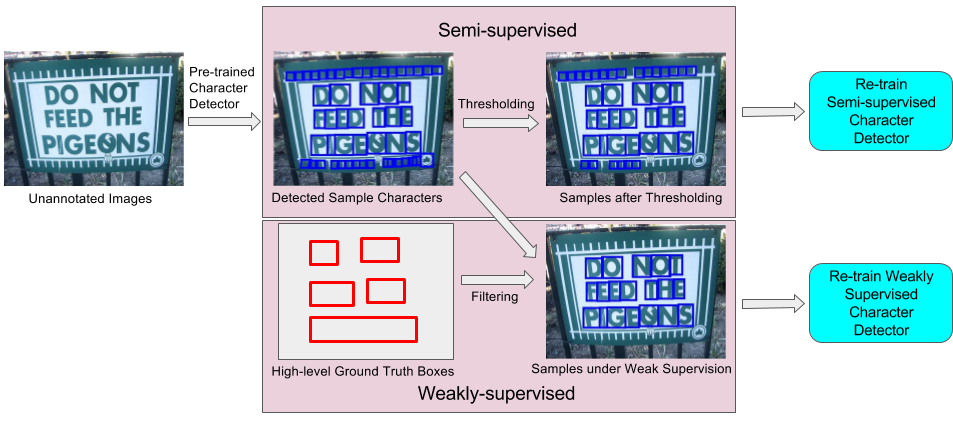} 
	\caption{The framework of the proposed WeText system: A ``light" supervised model is pre-trained using a small amount of annotated character image set. The light model is then applied to an unannotated dataset to search for more character samples which are combined with the small annotated dataset to train a semi-supervised model. Under certain weak annotations, better character samples can be searched to train a semi-supervised model.}
	\label{Fig:WeText}
\end{figure*}

\section{Related Work}
\label{sec:related_work}

Most existing text detection methods can be broadly classified into two categories, namely, character detection based and word detection based. The character detection based methods usually first detect multiple character candidates using various techniques, including sliding windows \cite{chen2004detecting, jaderberg2014deep, tian2015text}, MSERs \cite{he2016text, huang2014robust, neumann2011text,  neumann2012real, yin2014robust, zhang2016multi}, as well as some sophistically designed stroke detector \cite{epshtein2010detecting, huang2013text, yao2012detecting, zhang2015symmetry}. The detected character candidates are filtered by a text/non-text classifier to remove false candidates. Finally, the identified characters are grouped into words/text lines by either heuristic rules \cite{huang2014robust, yin2014robust, zhang2016multi} or sophisticated clustering/grouping models \cite{pan2011hybrid, tian2015text}. Though the initial character candidate detection can achieve very high recall, the current approach involving multiple sequential steps accumulates error which often degrades the final performance greatly. In particular, the intermediate text/non-text classification step requires a large amount of annotated character images which are very time consuming and costly to prepare.

The methods in the second category instead detect words directly \cite{gupta2016synthetic, he2016accurate, jaderberg2016reading, liao2016textboxes, polzounov2017wordfence, tian2016detecting, zhong2016deeptext}. In \cite{jaderberg2016reading}, object region proposals are employed to first detect multiple word candidates which are then filtered by a random forest classifier and the word bounding boxes are finally fine-tuned with Fast R-CNN \cite{girshick2015fast}. An Inception-RPN word proposal network \cite{zhong2016deeptext} is proposed which employs Faster R-CNN \cite{ren2015faster} to improve the word proposal accuracy. Gupta \etal \cite{gupta2016synthetic}  introduce a Fully-Convolutional Regression Network to jointly achieve text detection and bounding-box regression at multiple image scales. Tian \etal \cite{tian2016detecting} propose a Connectionist Text Proposal Network that combines CNN and long short-term memory (LSTM) architecture to detect text lines directly. The most recent TextBoxes approach \cite{liao2016textboxes} designs an end-to-end trainable network to output the final word boxes directly, exploiting state-of-the-art (SSD) object detector \cite{liu2016ssd}. Though the word detection approach is simpler, it does not work well with multi-oriented texts due to the constraints on word proposals. In addition, visually defining a word boundary may not be feasible for texts in many non-Latin languages such as Chinese.

Inspired by the idea of weakly supervised learning \cite{joulin2016learning, papandreou2015weakly}, we propose a weakly supervised scene text detection framework that learns on a small amount of character-level annotated text images, followed by boosting the perfomrance with a much larger amount of weakly annotated images at word/text line level. In the scene text reading domain, similar weakly supervised learning idea has been explored for scene text recognition problem \cite{bissacco2013photoocr, jaderberg2014deep}. In  \cite{bissacco2013photoocr}, a self-supervised training mechanism is designed to augment the training data. In particular, an initial recognition model trained with five million images is applied to search for new training samples where the alignment between images and text is utilized to enhance the quality (based on the assumption that text in real word images also exists verbatim on the web). In \cite{jaderberg2014deep}, similar idea is adopted for automated data mining of Flickr imagery that automatically generates word and character level annotations. The weak correspondence between texts in image titles and texts in scene images is utilized to search for positive training samples.

\section{The Proposed Method}
\label{sec:wetext}

\subsection{WeText Framework}
\label{sec:wetext_sub}

This section describes the system framework of the proposed weakly supervised scene text detection technique. The system consists of three components including unified scene character detection, semi-supervised and weakly supervised scene text modeling, and graph based text line extraction. The unified scene character detection aims to determine a bounding box together with a confidence score for each character in scene images. 
The semi-supervised and weakly supervised scene text modeling is achieved by learning from unannotated or weakly annotated scene text images automatically,  as illustrated in Figure \ref{Fig:WeText}. 
The graph based text line extraction algorithm \cite{tian2015text} is adopted to group characters into text lines.

\subsection{Unified Scene Character Detection}

We detect characters in scene images by exploiting the recent SSD framework \cite{liu2016ssd} which is designed for generic object detection and has demonstrated superior performance. The adoption of this regression based network aims to address the low efficiency and error accumulation issues of the current scene character detection paradigm where character detection and classification are designed as two separate processes. To the best of our knowledge, this is the first attempt that makes use of the regression based deep networks for scene character detection.

The early layers of the SSD character detector network are based on a standard architecture (VGG-16 \cite{simonyan2014very}) used for image classification. The last two fully-connected layers are converted into convolutional layers with subsampled parameters to speed up the computation. Two auxiliary structures are stacked to the base network to produce character predictions. First, additional convolutional layers are added to the end of the base network, allowing predictions at multiple scales. Unlike Faster R-CNN \cite{ren2015faster} which uses a single feature layer for prediction, SSD selects multiple feature layers including layers in base network and those additional stacked ones. 
Second, predictions are computed by applying a set of $3*3$ filters to each of the selected feature layers. At each location in the feature layer, we need to predict 6 values for each default anchor, i.e., 4 offsets of the bounding box and 2 scores (text/background).

\begin{figure}[!t]
	\centering		
	\includegraphics[width=0.49\linewidth, height=0.15\textheight]{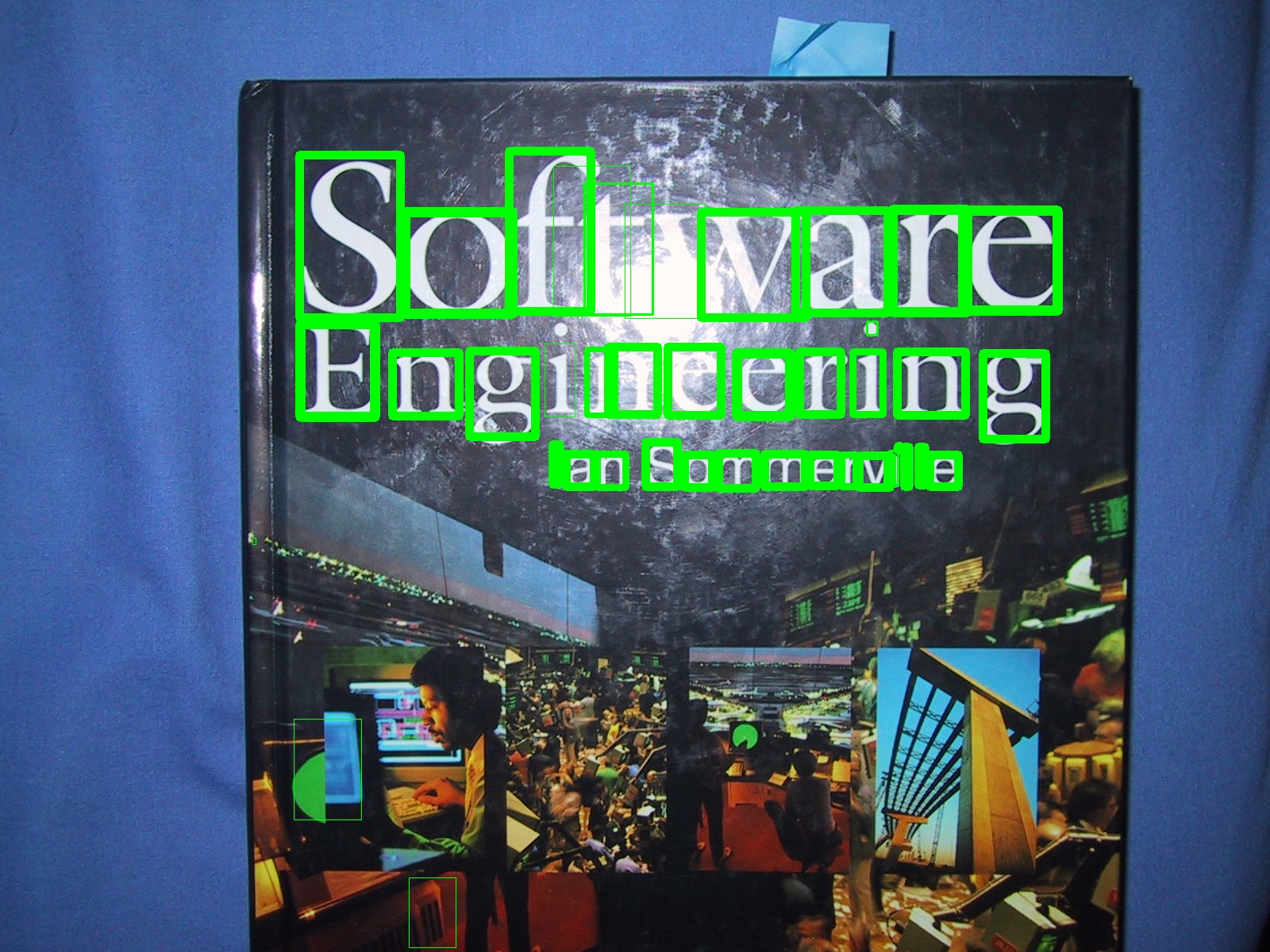} 
	\includegraphics[width=0.49\linewidth, height=0.15\textheight]{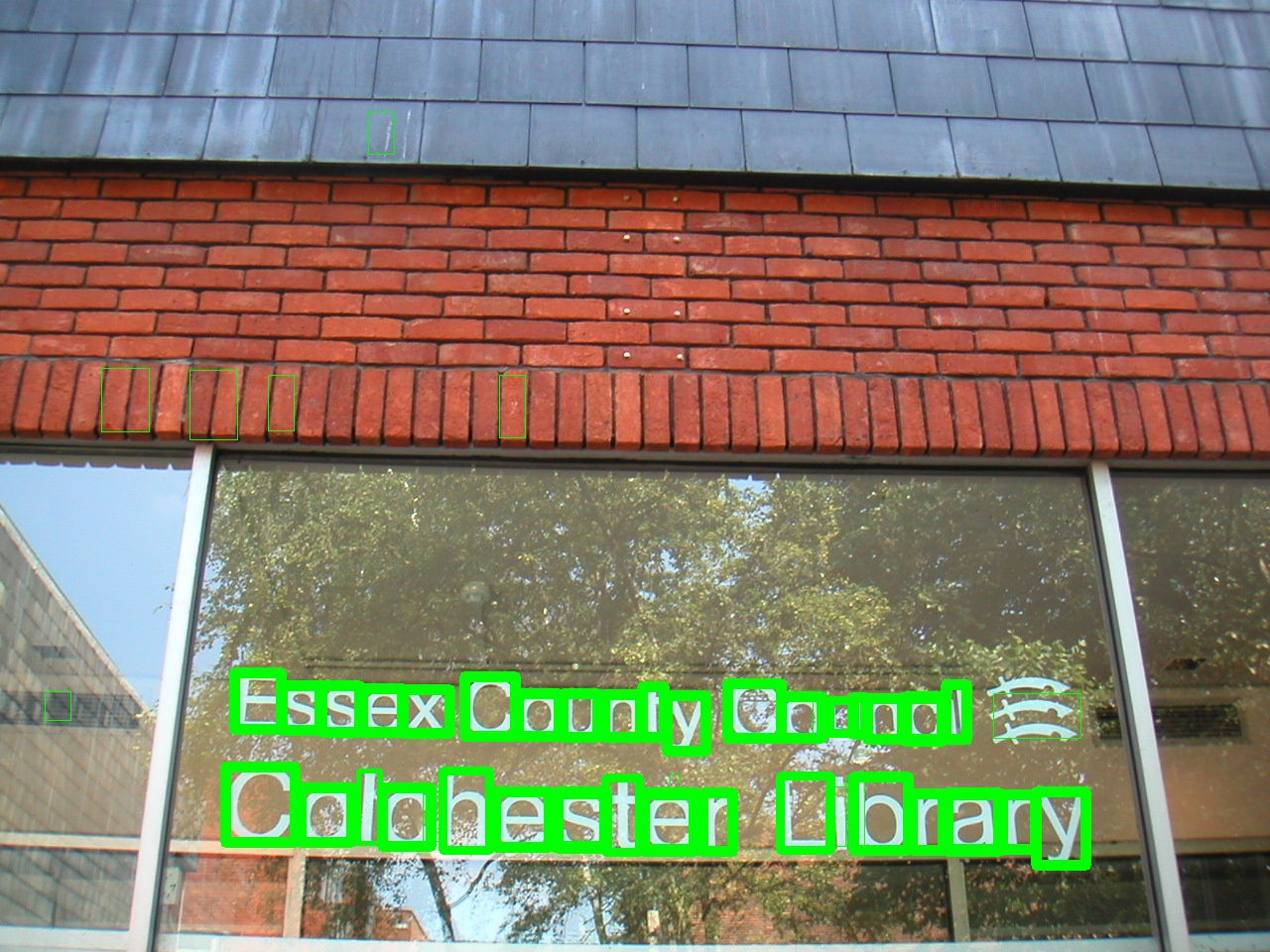}
	\caption{Character detection by the baseline model trained using the ICDAR 2013 training images. Thicker bounding boxes indicate higher detection confidence.}
	\label{Fig:ssd_char_output}
	\vspace{3mm}
\end{figure}

At the inference stage, Non-Maximum Suppression (NMS) is applied with Jaccard overlap of 0.45 to reduce the detection boxes. As characters tend to appear in groups, the average number of characters in each image is much larger than general objects in scenes. We therefore keep the top 1000 candidates before NMS and the top 500 detections after NMS per images instead of top 400 and top 200 as used in \cite{liu2016ssd}. Examples of the proposed character detection model are given in Figure \ref{Fig:ssd_char_output} (the thickness of the box boundary lines indicates the detection confidence).

\subsection{WeText Learning}

We investigate two learning strategies to deal with the limited annotation issue which widely exists in many other deep learning systems for object detection/recognition tasks. The first is semi-supervised learning that aims to exploit a large amount of completely unannotated text images. The second is weakly supervised learning where the text images are annotated at the word/text line (instead of character) level. Under both data scenarios, we assume that we have a ``light" supervised scene character detection model that is pre-trained by using a small amount of fully annotated scene text images. More details are to be described in the ensuing two subsections.

\subsubsection{Semi-Supervised Learning}
\label{sec:semi_sup}

The scenario of the semi-supervised learning here aims to improve a detection model by learning from a large amount of unannotated data $R$. Specifically, we have a scene text detection model $M$ that is pre-trained by using a small amount of fully annotated scene character images $D$, and a large amount of scene text images $R$ that completely has no annotations. The target is to improve $M$ by learning from $R$ with as less manual intervention as possible. It is actually a generic deep learning problem while facing various unannotated ``Big Data".

In the WeText system  as illustrated in Figure \ref{Fig:WeText}, we first run the pre-trained model $M$ on the unannotated dataset $R$. For each image in $R$, the model $M$ returns a set of candidate character bounding boxes as well as the corresponding detection score $C= \{ (c_1, s_1), (c_2, s_2) ..., (c_i, s_i), ...\}$. The positive character samples can be identified by a confidence threshold: 
\begin{equation}
\label{eq:semi_sup}
\begin{aligned}
P = \{ \: c_i  \: |  \: s_i > S \;\; and \;\; c_i \in C \: \}
\end{aligned}
\end{equation}
where $s_i$ denotes the detection score of the $i$-th detected character candidate $c_i$. The notation $S$ is the detection confidence threshold that is used to identify the positive samples. Note that $S$ cannot be too large otherwise the identified sample images lose diversity. At the same time, $S$ cannot be too small otherwise a large amount of non-text samples will be returned. Our experiments show that scene characters can be well searched when $S$ is set to around ${[\,0.4, \, 0.6 \,]}$.

The finally identified positive character sample set $P$ can then be combined with the annotated image set $D$ to train a more robust and accurate scene character detector $M'$. The top right and bottom left images in Figure \ref{Fig:ssd_char_detector_cmp} show the scene characters that are detected by $M$ and $M'$, respectively. It can be seen that the semi-supervised model $M'$ clearly outperforms the initial model $M$.

\subsubsection{Weakly Supervised Learning}
\label{sec:weak_sup}

\begin{figure}[!t]
	\centering	
	\includegraphics[width=0.4\linewidth, height=0.15\textheight]{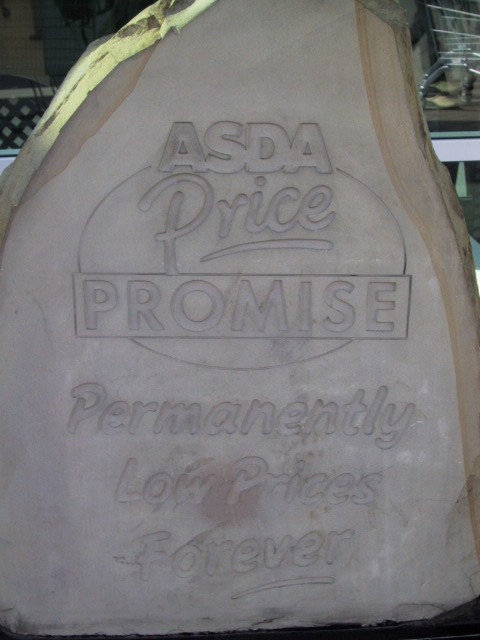} \hspace{1mm}
	\includegraphics[width=0.4\linewidth, height=0.15\textheight]{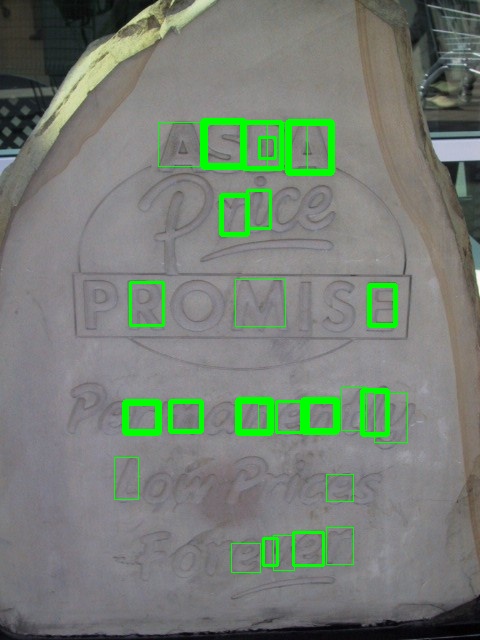}	
	\\ \vspace{1mm}
	\includegraphics[width=0.4\linewidth, height=0.15\textheight]{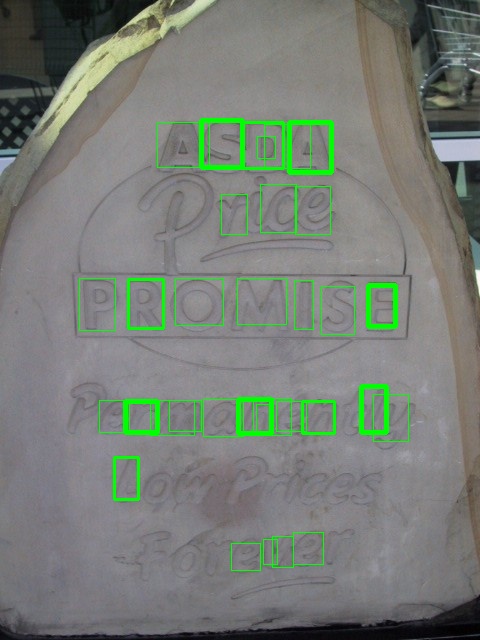} \hspace{1mm}	
	\includegraphics[width=0.4\linewidth, height=0.15\textheight]{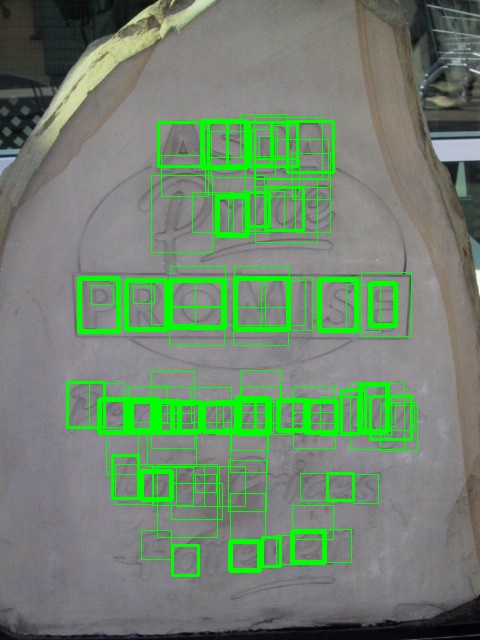}
	\caption{Comparison of different character detectors. Images in the top row from left to right are the input image and output of the baseline detector. Images in the bottom row from left to right are outputs of ``COCO-Text\_Semi"  and ``COCO-Text\_Weakly" detectors, respectively. The thickness of the box boundary lines indicates the detection confidence.}
	\label{Fig:ssd_char_detector_cmp}
	\vspace{3mm}
\end{figure}

The weakly supervised learning in WeText aims to improve a scene character detection model by learning from large amounts of weakly annotated text images. Different from the semi-supervised learning as described in the last subsection, we have a large dataset $R'$ that has weak annotations at word/text line level as denoted by a set of word/text line bounding boxes $G= \{g_1, g_2, ..., g_j, ...\}$. The target is to improve $M$ by learning from $R'$ with as less manual intervention as possible. Compared with the semi-supervised learning, the weakly supervised learning has high-level annotations of words/text lines which provide very useful guidance while searching for scene characters in $R'$.

Similar to the semi-supervised learning as described in the last subsection, the pre-trained model $M$ is first applied to the weakly annotated dataset $R'$, and a candidate character set $C$ is accordingly detected for each image within $R'$. With the weak annotation $G$ at word/text line level, the positive character sample images are determined as follows:
\begin{equation}
\label{eq:weak_sup}
\begin{aligned}
P' = \{ \; c_i  \: |  \: & s_i > S' \;\; and \;\; c_i \in C \: \\
& and \;\; I_{x_i} \; / \; W_{c_i} > T_x \\
& and \;\; I_{y_i} \; / \; H_{c_i} > T_y \; \}
\end{aligned}
\end{equation}
where $W_{c_i}$ and $H_{c_i}$ denote the width and height of the detected character candidate $c_i$, $I_{x_i}$ and $I_{y_i}$ denote the maximum horizontal and vertical intersection between $c_i$ and all ground truth bounding boxes in $G$. $S'$ is a predefined confidence threshold to select positive candidates. It can be set at a much lower value between $[0.2, 0.3]$ due to the constraint provided by the high-level annotations. $T_x$ and $T_y$ are both set at 0.8, based on the observation that a detected character candidate box with more than 80\% overlap with the ground truth word/text line boxes are usually texts. 

The identified positive sample image set $P'$ can then be combined with the annotated image set $D$ to train a more robust and accurate scene character detector $M''$. The bottom right image in Figure \ref{Fig:ssd_char_detector_cmp} shows the scene characters detected by $M''$. It can be seen that the weakly supervised detector $M''$ outperforms both the initial detector $M$ and the semi-supervised $M'$ clearly.

The better performance of the weakly supervised learning can be explained by two factors. First, more falsely detected character candidates can be removed by leveraging on the annotation bounding boxes at the word/text line level. Second, a lower text confidence threshold $S'$ can be set with the guidance of word/text line bounding boxes which helps to detect more positive character samples greatly. Therefore, the weakly supervised learning can search and retrieve more positive samples of higher quality as compared with the semi-supervised learning.

\section{Experiments}
\label{sec:experiments}

Our experiments involve four datasets including the ICDAR 2013 dataset \cite{karatzas2013icdar}, the FORU dataset \cite{zhong2016deeptext}, the COCO-Text dataset \cite{veit2016coco} and the SWT dataset \cite{epshtein2010detecting}.

\subsection{Datasets}

\textbf{ICDAR 2013}
\noindent
\footnote{\url{http://rrc.cvc.uab.es/?ch=2&com=downloads}} consists 229 training image and 233 testing images. Each image also has a segmentation map which helps to extract character boxes. In the experiments, the 229 training images with character-level boxes are used to pre-train a baseline character detector, and the 233 testing images are used for evaluation following the protocol in \cite{karatzas2013icdar}.

\textbf{FORU}
\noindent
\footnote{\url{https://pan.baidu.com/s/1kVRIpd9}} is collected from the Flickr website. In our experiments, we use the English2k sub-dataset with 1162 images and 14888 annotated characters. Both character-level and word-level bounding boxes will be used to evaluate the proposed weakly supervised learning.

\textbf{COCO-Text}
\noindent
\footnote{\url{http://vision.cornell.edu/se3/coco-text/}} is derived from the MS COCO dataset with ``incidental" texts. In our experiments, we use the training images with at least one legible English text region which leads to 14712 images. Note only word-level bounding boxes are annotated on this dataset.

\textbf{SWT}
\noindent is introduced in \cite{epshtein2010detecting} which contains 307 images with word bounding boxes for testing. The dataset is very challenging with cluttered scene images under low contrast and it also contains many small text regions. For evaluation, we use the protocols provided by the dataset creators.

\subsection{Implementation Details}

Similar to the original SSD \cite{liu2016ssd}, we also fine tune from the pre-trained VGG-16 network \cite{simonyan2014very} with initial learning rate $10^{-3}$, momentum $0.9$, weight decay $5*10^{-4}$ and batch size $32$ for all the experiments. In addition, all character detection models are trained with input of image scale $512*512$ and tested at a \textbf{single} image scale $600*600$. Further, the parameters in  Equation \ref{eq:semi_sup} and \ref{eq:weak_sup} are empirically set at $S$ = $0.5$, and $S'$ = $0.2$ for all experiments. The text confidence threshold for the weakly supervised learning is much lower than that for the semi-supervised learning because word-level bounding boxes in the weakly supervised learning helps to better remove false positives and retrieve true positive samples with lower scores.

The initial ``light" character detection model is trained using character annotations within the 229 training images in the ICDAR 2013 dataset. 15k learning iterations are set and the learning rate is reduced to $10^{-4}$ after 10k iterations. This model will serve as the \textbf{Baseline} for both character detection and text line detection as shown in Tables \ref{tab:icdar13_char}. Experiments on the FORU dataset and the COCO-Text dataset target to improve this Baseline model by deriving more positive training samples  from the two datasets.

\vspace{1mm}
\textbf{FORU}
\noindent
We study three settings on this dataset including \textbf{1)} Fully supervised learning where the ground truth character bounding boxes of this dataset are directly combined with the ICDAR training character images to train a better model. This experiment sets an upper bound for the usage of the FORU dataset and experimental result will be used to verify the effectiveness of the semi-supervised and weakly supervised learning;\textbf{ 2)} Semi-supervised learning where no annotation information is used and  positive samples are obtained as described in Section \ref{sec:semi_sup}; and \textbf{3)} Weakly supervised learning where ground truth word bounding boxes are used to guide the sample image searching as described in Section \ref{sec:weak_sup}. For all the three settings, we fine-tune from the initial character detector for 3k iterations with learning rate $10^{-3}$ which is reduced $10^{-4}$ for another 1k iterations and further reduced to $10^{-5}$ for the last 1k iterations.

\vspace{1mm}
\textbf{COCO-Text}
\noindent
We only study the semi-supervised and weakly supervised settings for this dataset as it does not have character-level bounding boxes. Similar to the FORU dataset, we fine-tune from the character detector trained on the ICDAR 2013 for 10k iterations with learning rate $10^{-3}$ which is reduced $10^{-4}$ for another 3k iterations and further reduced to $10^{-5}$ for the last 2k iterations.

For each setting on the FORU and COCO-Text datasets, the derived positive training samples are combined with the initial ICDAR 2013 training images to re-train a character detection model. Hence we have another five character detection models including \textbf{FORU\_GT}, \textbf{FORU\_Weakly}, \textbf{FORU\_Semi}, \textbf{COCO-Text\_Semi}, and \textbf{COCO-Text\_Weakly} as listed in Table \ref{tab:icdar13_char}. Leveraging on the six character detection models (five newly trained plus the Baseline model), we have six corresponding text line detection models after incorporating text line extraction process, including \textbf{Baseline\_TL}, \textbf{FORU\_GT\_TL}, \textbf{FORU\_Semi\_TL}, \textbf{FORU\_Weakly\_TL}, \textbf{COCO-Text\_Semi\_TL}, and \textbf{COCO-Text\_Weakly\_TL} as listed in Table \ref{tab:icdar13_res}.

\subsection{Experimental Results}

We evaluate the WeText framework on the 
ICDAR 2013 testing dataset and the SWT dataset. 

\subsubsection{Character Detection}

\begin{figure}[!t]
	\centering		
	\includegraphics[width=0.7\linewidth, height=0.2\textheight]{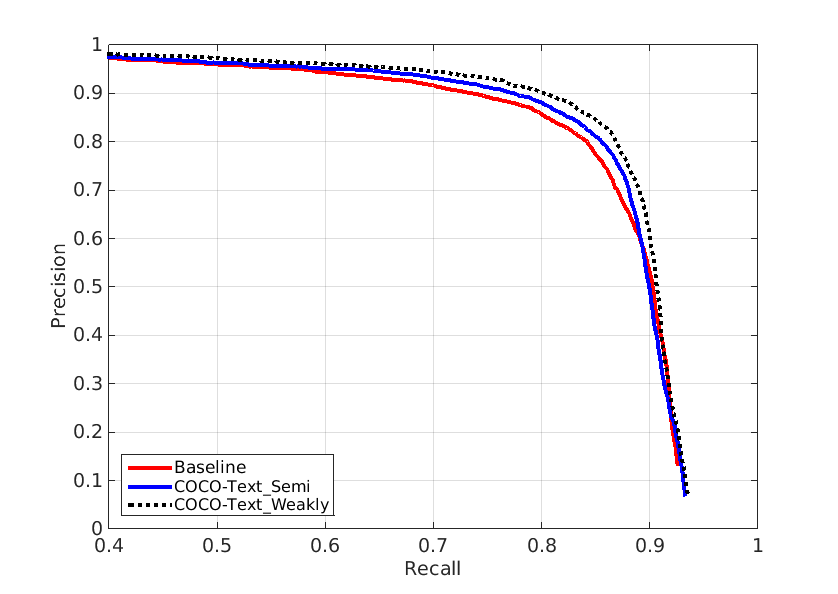} 
	\caption{Comparison of character detection performance on ICDAR 2013 test dataset under different learning schemes from COCO-Text dataset. Baseline detector is trained only on ICDAR 2013 training dataset with character boxes. ``COCO-Text\_Semi" and ``COCO-Text\_Weakly" detectors are trained without annotation and with text block bounding boxes as described in Section \ref{sec:semi_sup} and Section \ref{sec:weak_sup}, respectively. }
	\label{Fig:char_pr_curve}
\end{figure}

We first show the character detection performance on the ICDAR 2013 test dataset, to validate the effectiveness of the proposed semi-supervised and weakly supervised learning from COCO-Text dataset. The PASCAL VOC \cite{everingham2010pascal} intersection-over-union (IoU) overlap is used as the evaluation metric (positive detection if IoU$\geq0.5$). The recall-precision curve in Figure \ref{Fig:char_pr_curve} shows that the semi-supervised model performs clearly better than the baseline model. In addition, the weakly supervised model is superior to both the baseline model and the semi-supervised model. The remarkable performance is largely due to the high precision where the word/text line level ground truth boxes help to filter out lots of false positive samples.

Table \ref{tab:icdar13_char} shows the precision, recall, and F-score of all character detection models described in the previous subsection, where a confidence threshold $0.05$ is used for all detected character candidates (on the ICDAR 2013 testing images). The confidence threshold $0.05$ is used for the optimal text line extraction to be described in the next subsection. As Table \ref{tab:icdar13_char} shows, both semi-supervised and weakly supervised models obviously surpass the baseline model. At the same time, the weakly supervised model clearly outperforms the semi-supervised model due to the availability of the high-level annotations.

\begin{table}[!t]
	\centering
	\small
	\renewcommand{\arraystretch}{1.1}
	\vspace{5mm}
	\caption{\footnotesize {\textbf{Character detection} results on ICDAR 2013 dataset (\%)} }
	\label{tab:icdar13_char}
	\vspace{-1mm}
	\scalebox{0.93} {
		\begin{tabular}{|l|c|c|c|}
			\hline
			Method 										   &Recall &Precision & F-score \\ 		
			\hline
			\textbf{Baseline} 								& 84.80 	& 61.44		& 71.26		\\ 	
			\hline
			\textbf{FORU\_Semi} 							& 85.71		& 63.65		& 73.05		\\ 	
			\hline
			\textbf{FORU\_Weakly} 							& 85.18 	& 67.59		& 75.37 		\\ 	
			\hline
			\textbf{FORU\_GT} 								& 85.37 	& 71.83		& 78.02 		\\ 
			\hline
			\textbf{COCO-Text\_Semi} 						& 85.35		& 66.74		& 74.91 		\\ 	
			\hline
			\textbf{COCO-Text\_Weakly} 						& 85.45		& 72.39		& 78.38 		\\ 	
			\hline
		\end{tabular}
	}
\end{table}

\subsubsection{Text Line Extraction}

The detected characters are grouped into text lines using the TextFlow algorithm \cite{tian2015text}, where we use all detected character candidates that have a detection confidence larger than $0.05$. The use of a much smaller confidence threshold (as compared with the $S$ and $S'$ that are used for semi-supervised and weakly supervised training) is because the min-cost flow based text line extraction helps to remove lots of false positive character candidates.


\begin{table}[!t]
	\centering
	\small
	\renewcommand{\arraystretch}{1.1}
	\caption{\small {\textbf{Text Line detection} results on ICDAR 2013 dataset (\%)} }
	\label{tab:icdar13_res}
	\vspace{-1mm}
	\scalebox{0.87} {
		\begin{tabular}{|l|c|c|c|c|}
			\hline
			Method 													& Year 	&Recall &Precision & F-score \\ 		
			\hline
			Lu \etal \cite{lu2015scene} 							& 2015 	& 69.6	& 89.2	& 78.2  \\	
			\hline
			Tian \etal \cite{tian2015text} 							& 2015 	& 75.9	& 85.2	& 80.3  \\
			\hline	
			Liao \etal \cite{liao2016textboxes} (single scale) 		& 2017 & 74.0	& 88.0	& 81.0 	\\	
			\hline											
			Zhang \etal \cite{zhang2016multi} 						& 2016 & 78.0	& 88.0	& 83.0 	\\								
			\hline
			Gupta \etal \cite{gupta2016synthetic} 					& 2016 & 75.5	& \textbf{92.0}	& 83.0	\\
			\hline
			Liao \etal \cite{liao2016textboxes}  (multi-scale)		& 2017 & 83.0	& 89.0	& 86.0 \\
			\hline
			He \etal \cite{he2016accurate} 							& 2016 & 83.0	& 90.0	& 86.0 \\  																					
			\hline
			\textbf{Baseline\_TL} 								&-	& 80.7	& 84.2	& 82.3 		\\ 	
			\hline
			\textbf{FORU\_Semi\_TL} 							&- 	& 82.0	& 84.7	& 83.4 		\\ 	
			\hline
			\textbf{FORU\_Weakly\_TL} 							&- 	& 82.4	& 88.6	& 85.4 		\\ 	 	
			\hline
			\textbf{FORU\_GT\_TL} 								&- 	& 82.2	& 90.9	& 86.3 		\\ 	
			\hline
			\textbf{COCO-Text\_Semi\_TL} 						&- 	& 81.8	& 86.9	& 84.2 		\\ 	
			\hline
			\textbf{COCO-Text\_Weakly\_TL} 						&- 	& \textbf{83.1}	& 91.1	& \textbf{86.9} 		\\ 	
			\hline
		\end{tabular}
	}
\end{table}

\textbf{Quantitative Results}
\noindent
As shown in Table \ref{tab:icdar13_res}, we achieve state-of-the-art performance on the ICDAR 2013 dataset through the proposed weakly supervised learning strategy. As all our experiments are run at a single scale image, our method outperforms the method in \cite{liao2016textboxes} significantly by 6\% F-score (86.9\% vs 81.0\%) when the method in \cite{liao2016textboxes} also uses a single scale image as input. In fact, our method still perform better by 1\% than \cite{liao2016textboxes} where multi-scale testing are adopted. Besides, our baseline model even outperforms the model in Tian et al. \cite{tian2015text}. This verifies that the proposed character detector is much more accurate and robust considering the two methods both used similar min-cost flow based text line extraction algorithm.

In addition, models by all three learning schemes, i.e. semi-supervised, weakly supervised, and fully supervised, perform better than the baseline model. In particular, the semi-supervised model improves more than 1\% and the fully supervised model achieves the best improvement by 4\%. The performance of the weakly supervised model is close to that of the fully supervised model, demonstrating the effectiveness of the proposed weakly learning scheme.

Furthermore, the text line extraction performance is further improved to 84.2\% and 86.9\%, respectively, for semi-supervised and weakly supervised models when the COCO-Text dataset is used. Similar to the FORU dataset, the ``COCO-Text\_Weakly\_TL" performs better than the ``COCO-Text\_Semi\_TL" for both recall and precision. This verifies that weakly labeled data can effectively helps to remove falsely detected samples and retrieve more difficult positive samples. Additionally, both ``COCO-Text\_Semi\_TL" and ``COCO-Text\_Weakly\_TL" outperform the ``FORU\_Semi\_TL" and ``FORU\_Weakly\_TL", respectively, demonstrating that a larger unannotated or weakly annotated dataset helps to train better semi-supervised and weakly supervised models.

To further verify the proposed framework, we also report results on the SWT dataset \cite{epshtein2010detecting} in Table \ref{tab:ms_swt_res}. All the settings are kept the same as those on ICDAR 2013 dataset except that the test scale is set to $800*800$. It can be seen that similar improvement is achieved as on the SWT dataset. The proposed method surpasses the baseline clearly and the learning from a bigger COCO-Text dataset outperforms the learning from a smaller FORU dataset. 

\begin{figure*}[!t]
	\centering	
	\includegraphics[width=0.22\linewidth, height=0.15\textheight]{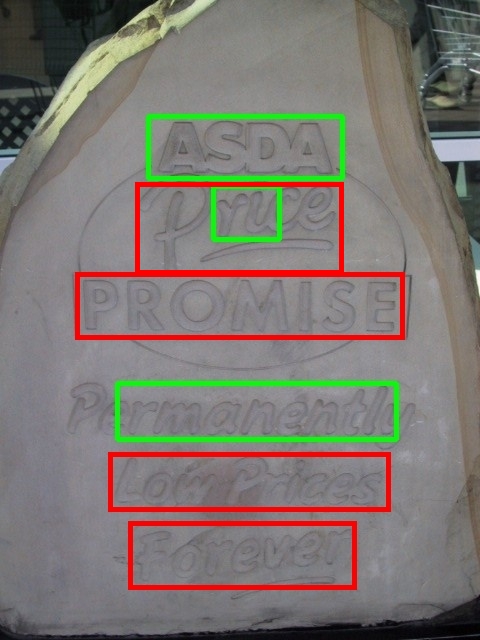}	
	\includegraphics[width=0.22\linewidth, height=0.15\textheight]{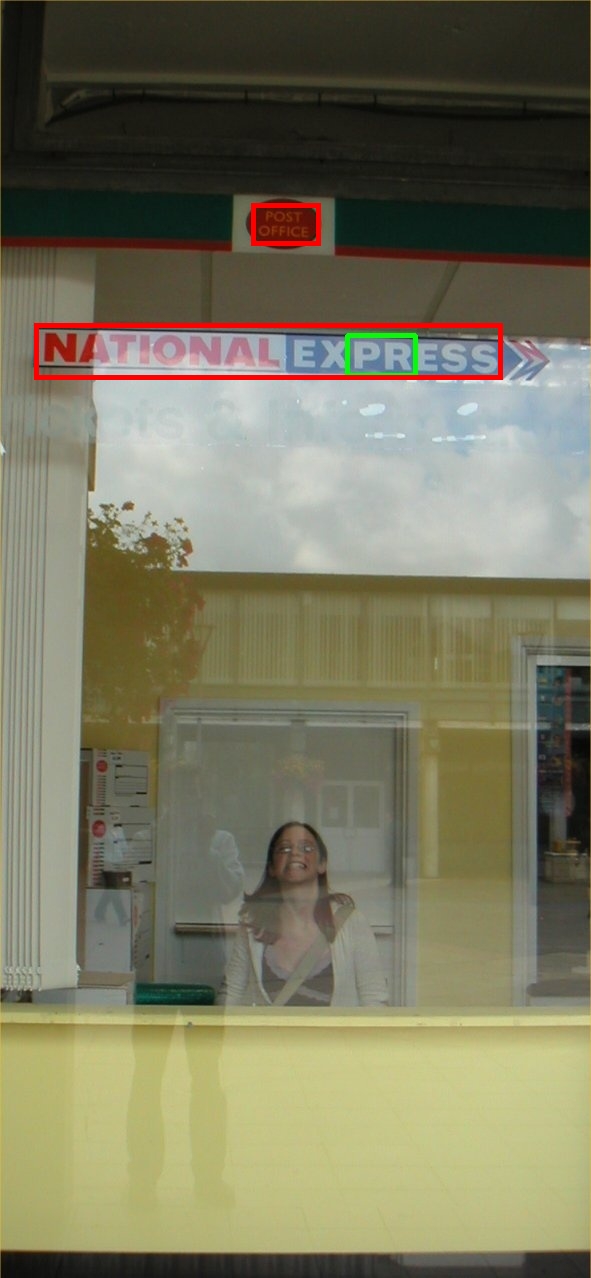}	
	\includegraphics[width=0.22\linewidth, height=0.15\textheight]{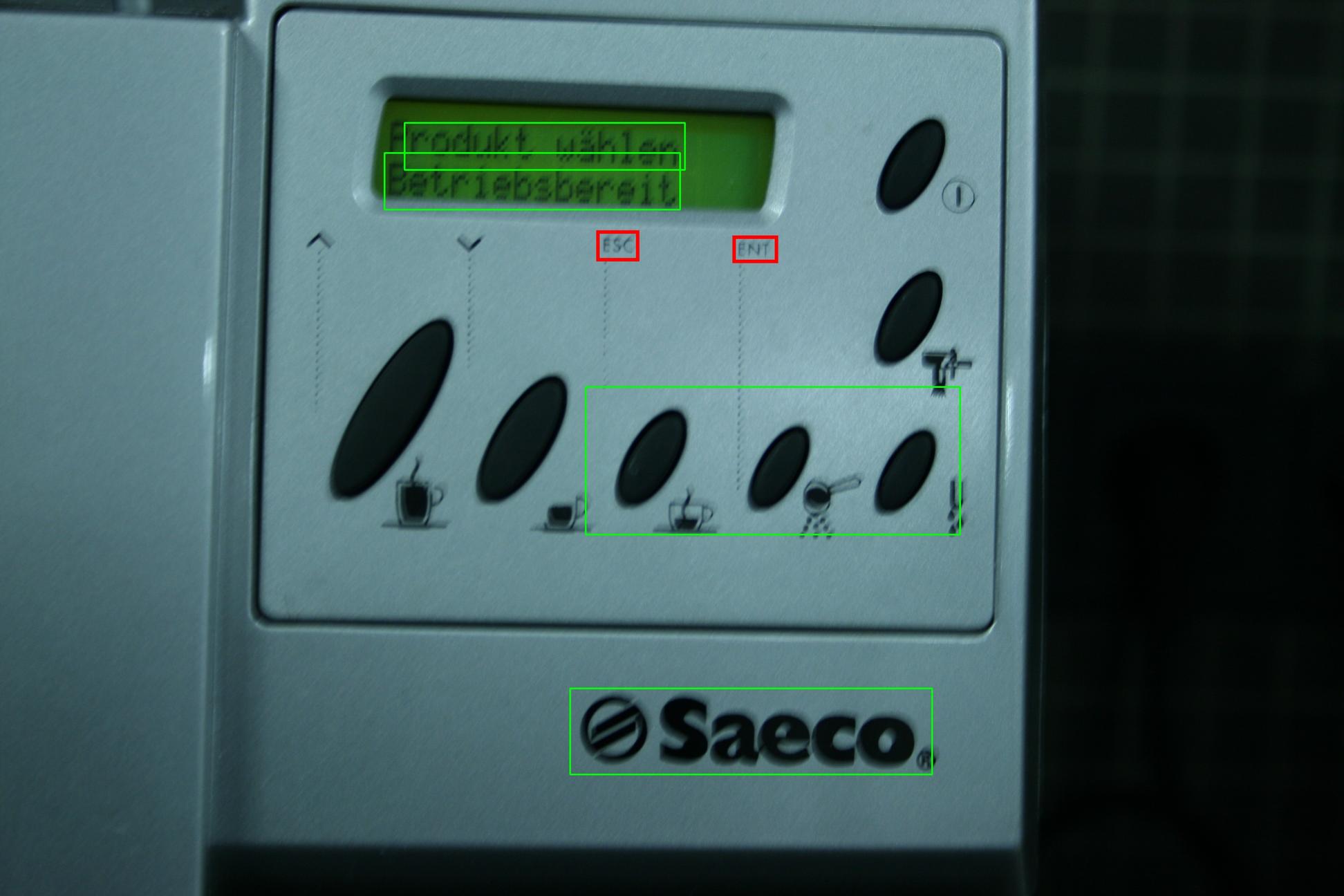}	
	\includegraphics[width=0.22\linewidth, height=0.15\textheight]{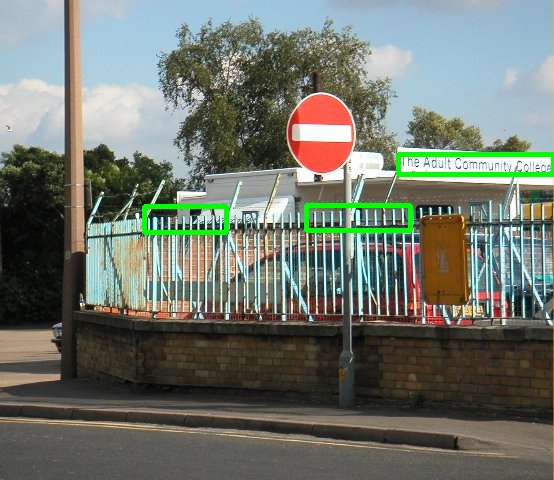}	
	\\ \vspace{1mm}
	\includegraphics[width=0.22\linewidth, height=0.15\textheight]{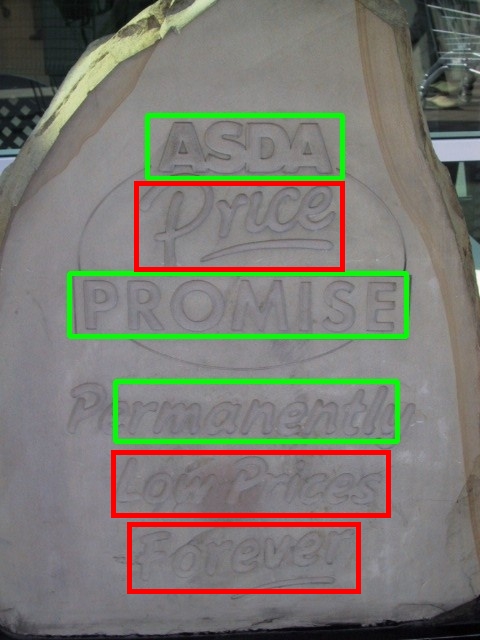}
	\includegraphics[width=0.22\linewidth, height=0.15\textheight]{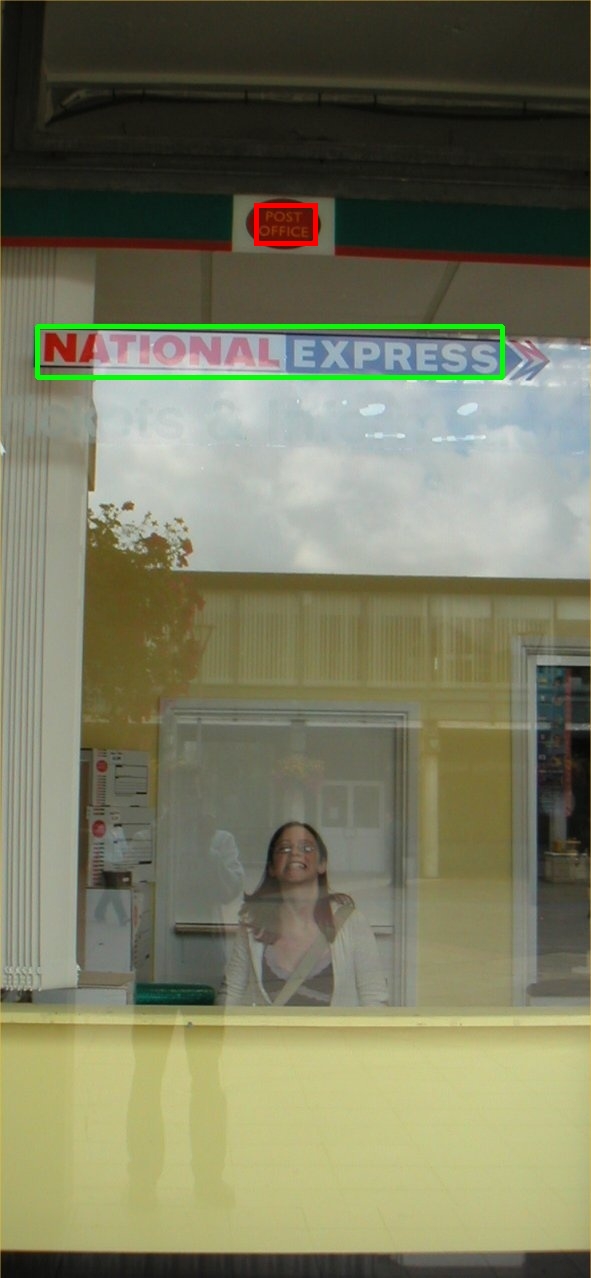}
	\includegraphics[width=0.22\linewidth, height=0.15\textheight]{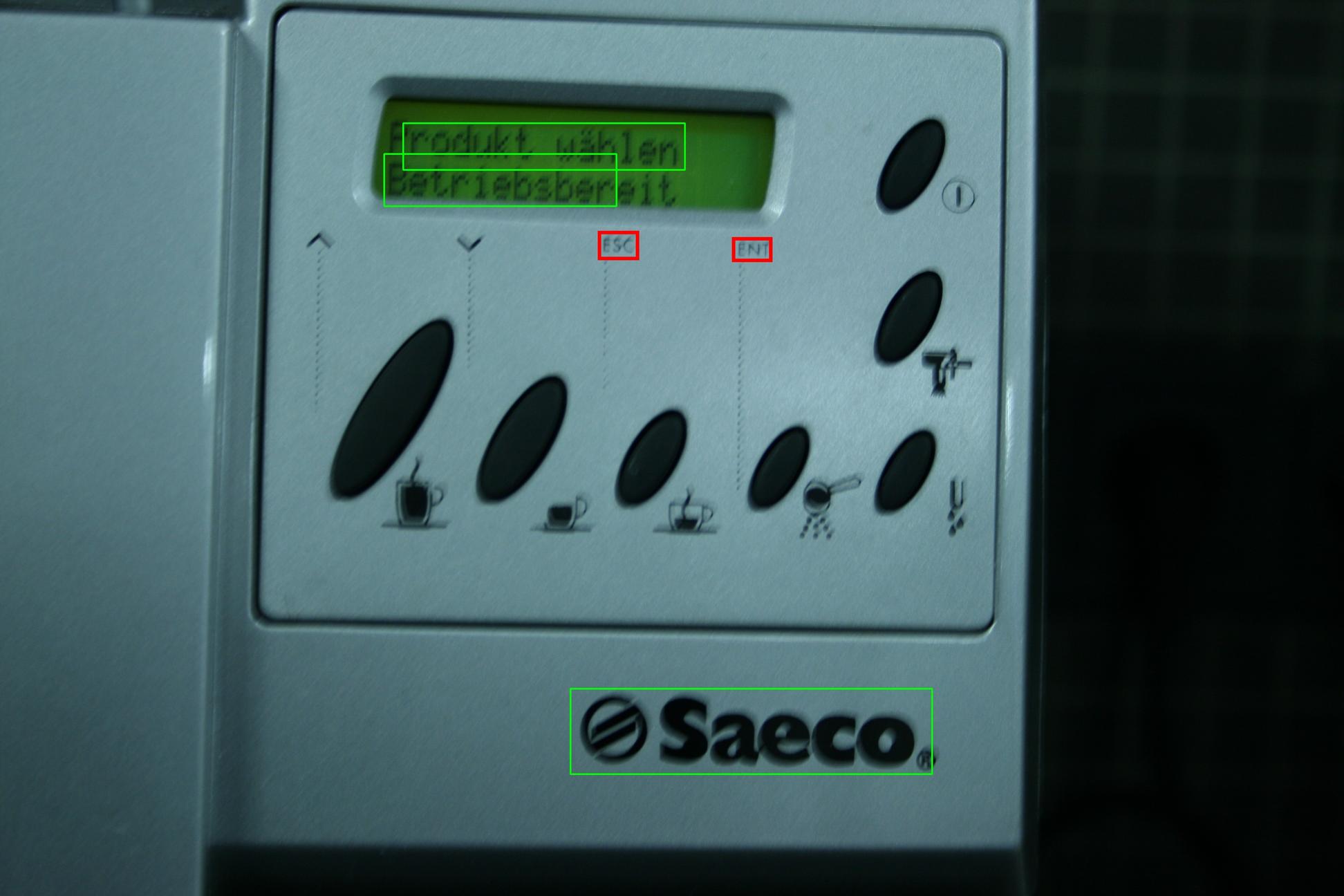}
	\includegraphics[width=0.22\linewidth, height=0.15\textheight]{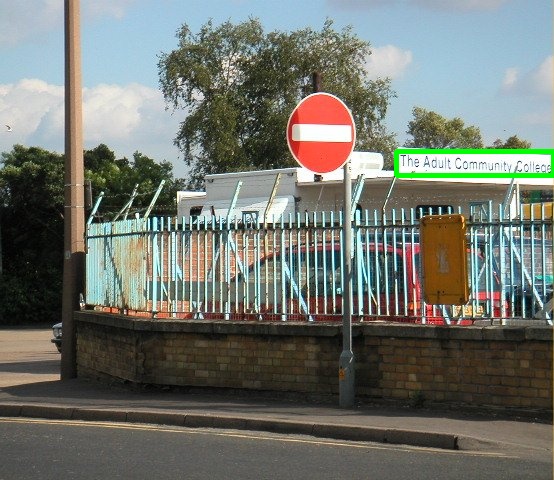}
	\\	\vspace{1mm}
	\includegraphics[width=0.22\linewidth, height=0.15\textheight]{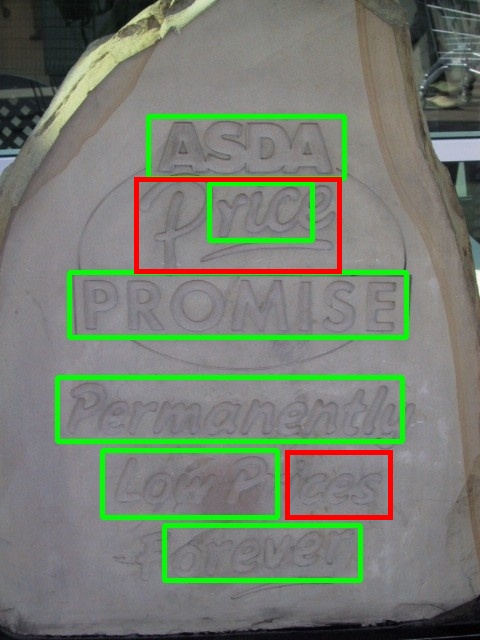}
	\includegraphics[width=0.22\linewidth, height=0.15\textheight]{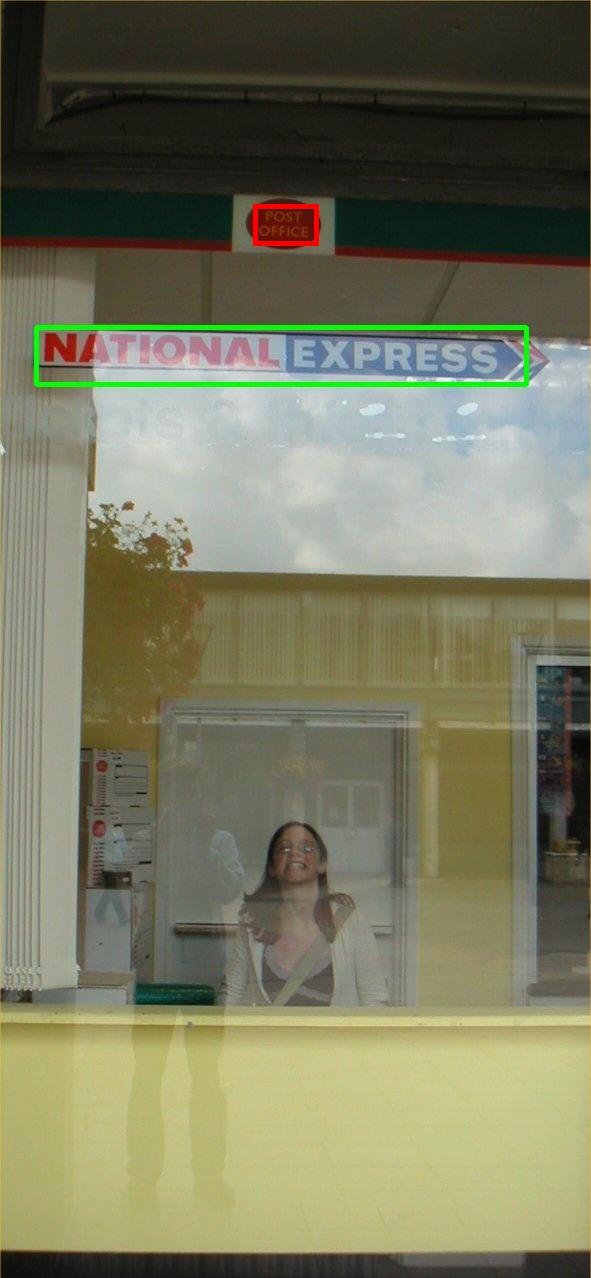}
	\includegraphics[width=0.22\linewidth, height=0.15\textheight]{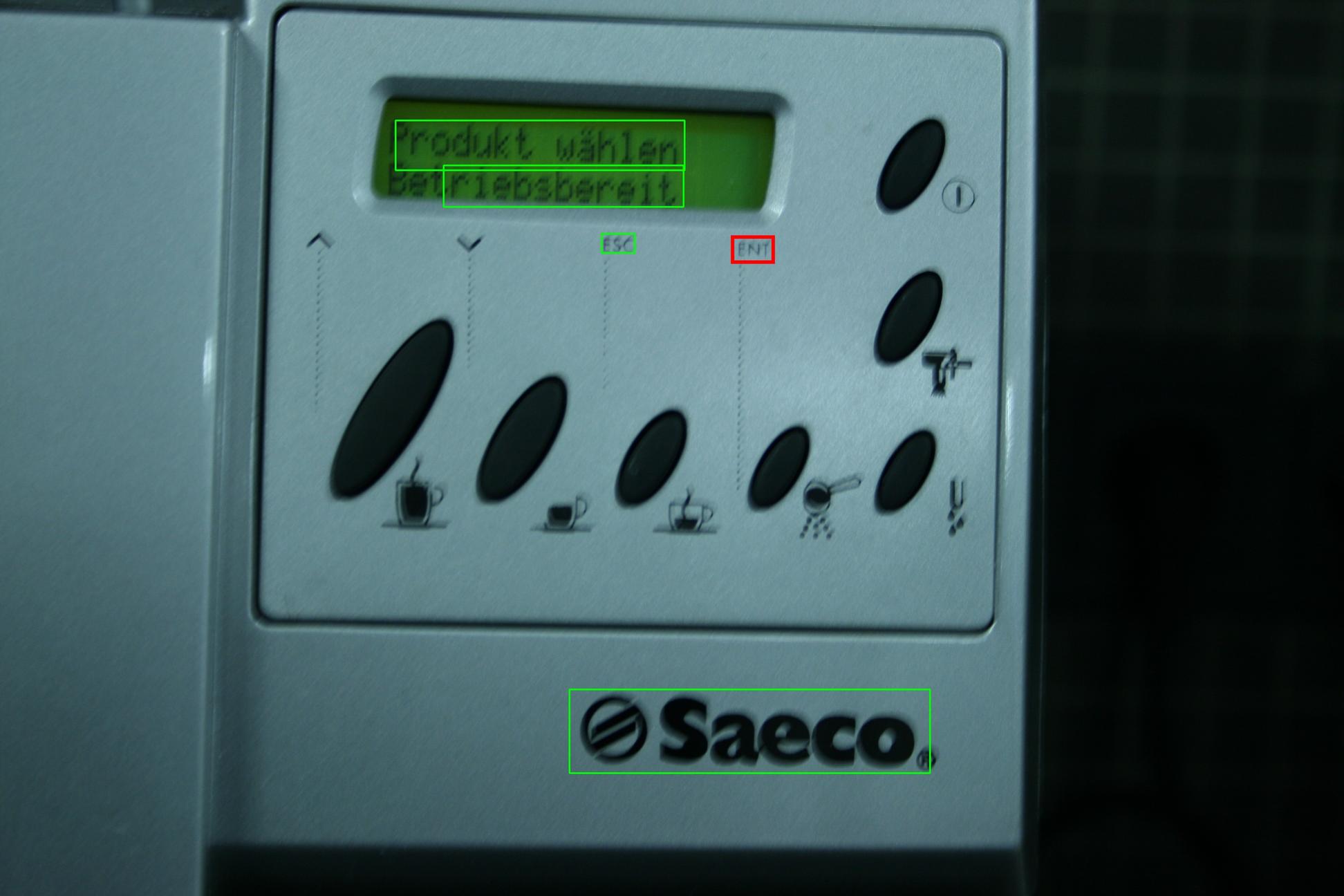}
	\includegraphics[width=0.22\linewidth, height=0.15\textheight]{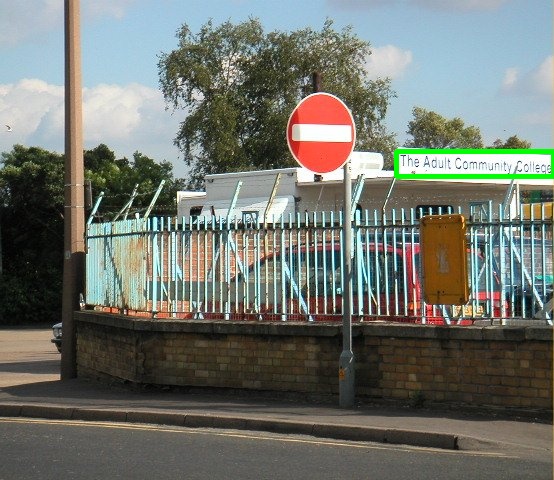}
	\\
	\caption{Comparison of text detection approach. Images from top to bottom are the text extraction outputs of the ``Baseline", ``FORU\_Weakly"  and ``COCO-Text\_Weakly" character detectors, respectively. Green boxes are outputs of our methods and red boxes are missing detections.}
	\label{Fig:ssd_line_comp}
	\vspace{3mm}
\end{figure*}

\textbf{Qualitative Results}
\noindent
Figure \ref{Fig:ssd_line_comp} shows text line extraction of several ICDAR 2013 test images that are processed by using the Baseline model, the ``FORU\_Weakly\_LT" model, and the ``COCO-Text\_Weakly\_TL" model, respectively. As Figure \ref{Fig:ssd_line_comp} shows, the scene text detection performance is clearly improved when more training samples are incorporated in the weakly supervised models. In particular, the recall of the first two sample images is greatly improved. False alarms are successfully removed in the third and forth images. In addition, the ``COCO-Text\_Weakly\_LT" detects one more small word than the ``FORU\_Weakly\_LT", demonstrating the advantage of learning from a much larger dataset. Overall, the proposed weakly supervised learning helps not only detect more positive texts but also remove more false alarms. On the other hand, it could still fail while handling handwriting texts, ultra-low contrast texts, etc. largely due to the limited amount of unannotated or weakly annotated text images. Some of the miss detections are marked by red bounding boxes in Figure \ref{Fig:ssd_line_comp}.

\begin{table}[!t]
	\centering
	\small
	\renewcommand{\arraystretch}{1.1}
	\caption{\footnotesize {\textbf{Text Line detection} results on SWT dataset (\%)} }
	\label{tab:ms_swt_res}
	\vspace{-1mm}
	\scalebox{0.93} {
		\begin{tabular}{|l|c|c|c|}
			\hline
			Method 										   &Recall &Precision & F-score \\ 	
			\hline
			Epshtein \etal \cite{epshtein2010detecting}		& 42.0 		& 54.0		& 47.0		\\	
			\hline
			Mao \etal \cite{mao2013scale}					& \textbf{58.0}		& 41.0		& 48.0		\\
			\hline
			Zhang \etal \cite{zhang2015symmetry}			& 53.0		& 68.0		& \textbf{60.0}		\\	
			\hline
			\textbf{Baseline} 								& 44.2 	& 69.0		& 53.9		\\ 	
			\hline
			\textbf{FORU\_Semi} 							& 47.5		& 68.8		& 56.2		\\ 	
			\hline
			\textbf{FORU\_Weakly} 							& 49.3 	& 67.9		& 57.1 		\\ 	
			\hline
			\textbf{FORU\_GT} 								& 48.2 	& \textbf{75.7}		& 58.9 		\\ 
			\hline
			\textbf{COCO-Text\_Semi} 						& 48.7		& 72.9		& 58.4 		\\ 	
			\hline
			\textbf{COCO-Text\_Weakly} 						& 49.7		& 74.9		& 59.8 		\\ 	
			\hline
		\end{tabular}
	}
\end{table}

\subsection{Discussion}

We also perform some preliminary study on iterative implementation of the proposed semi-supervised and weakly supervised learning schemes as described in Section \ref{sec:wetext}. Specifically, we repeat the positive sample searching and model re-training process by re-applying the newly trained character detection models back to the unannotated and weakly annotated dataset to search for more sample images for further model re-training. We evaluate the iterative learning idea on the FORU dataset. In the second round, the performance of the newly trained models ``FORU\_Semi\_TL" and ``FORU\_Weakly\_TL" improves from 83.4\% to 84.3\% and 85.4\% to 86.2\%, respectively, as compared with the re-trained models after the first round semi-/weakly supervised learning. In particular, the weakly supervised model ``FORU\_Weakly\_TL" after the second round performs nearly as good as the fully supervised model ``FORU\_GT\_LT". We also tested the models after the third round iterative learning  but little further improvement is observed. It is probably due to the very close performance to the fully supervised model and further improvements could be achieved when more unannotated or weakly annotated data become available.

The proposed technique is also fast. For the ICDAR 2013 test dataset, the proposed character detection model takes 0.19s per image and the text line extraction takes about 0.13s per image on Titan X GPU. The total processing time is about 0.32s on average which shows very good potential for various real-time scene text reading tasks.

\section{Conclusion}
\label{sec:conclusion}
In this paper, we propose a novel weakly supervised learning technique that aims to address the data annotation constraints which exist widely in most deep learning systems. Leveraging on a ``light" supervised model that is trained using a small amount of fully annotated images, two learning schemes, namely, semi-supervised learning and weakly supervised learning, are investigated by learning from a large amount of unannotated and weakly annotated images. The proposed technique is evaluated on two publicly available scene text datasets and experiments show that both semi-supervised and weakly supervised models outperform the ``light" supervised model clearly. In addition, the weakly supervised model performs almost as well as the fully supervised model.

{\small
\bibliographystyle{ieee}
\bibliography{egbib}
}

\end{document}